\documentclass[11pt,a4paper]{article}

\usepackage{graphicx}
\usepackage[cmex10]{amsmath}
\usepackage{amsfonts}
\usepackage[font=small]{caption}
\usepackage[margin=1in]{geometry}
\usepackage{fancyhdr}
\usepackage{color}
\usepackage{amssymb}
\usepackage{algorithm}
\usepackage{mathtools}
\usepackage[noend]{algpseudocode}
\usepackage{gensymb}
\usepackage[titletoc,title]{appendix}
\usepackage[noadjust]{cite}
\usepackage{epstopdf}
\usepackage{multirow}
\def\bb{\mathbf{b}}
\def\bc{\mathbf{c}}
\def\bd{\mathbf{d}}
\def\be{\mathbf{e}}

\def\bk{\mathbf{k}}

\def\br{\mathbf{r}}
\def\bs{\mathbf{s}}
\def\bt{\mathbf{t}}

\def\bx{\mathbf{x}}
\def\by{\mathbf{y}}

\def\sA{\mathsf{A}}

\def\sC{\mathsf{C}}
\def\sD{\mathsf{D}}

\def\sF{\mathsf{F}}

\def\sI{\mathsf{I}}

\def\sK{\mathsf{K}}

\def\sM{\mathsf{M}}
\def\sN{\mathsf{N}}

\def\sR{\mathsf{R}}
\def\sS{\mathsf{S}}

\def\sW{\mathsf{W}}

\newcommand{\norm}[1]{\left\lVert#1\right\rVert}

\def\uNc{\mathrm{N_c}}
\def\NoNumber#1{{\def\alglinenumber##1{}\State #1}\addtocounter{ALG@line}{-1}}

\usepackage{etoolbox}
\makeatletter
\patchcmd{\maketitle}
 {\def\@makefnmark}
 {\def\@makefnmark{}\def\useless@macro}
 {}{}
\makeatother

\begin{document}
%
%
\begin{center}
\textbf{Accelerating CS in Parallel Imaging Reconstructions Using an Efficient and Effective Circulant Preconditioner}
\end{center}
\textbf{Author list:} Kirsten Koolstra$^\dagger$$^1$, Jeroen van Gemert$^\dagger$$^2$, Peter~B\"ornert$^{1,3}$, \\ \bigskip Andrew~Webb$^1$, and~Rob~Remis$^2$.\\
$^\dagger$ These authors contributed equally to this work
\newline\\
\textbf{Corresponding author:}\\
Jeroen van Gemert, MSc\\
Circuits \& Systems Group of the Electrical Engineering, Mathematics and Computer Science faculty of the Delft University of Technology.\\
Mekelweg 4, 2628 CD Delft, The Netherlands \\
Email: J.H.F.vanGemert-1@tudelft.nl \\

\textbf{Word count:} 4839 \\

\textbf{Institution information:}\\
$^1$C. J. Gorter Center for High Field MRI\\
Department of Radiology\\
Leiden University Medical Center\\
Leiden, The Netherlands\\
\\
$^2$Circuits \& Systems Group\\
Electrical Engineering, Mathematics and Computer Science Faculty\\
Delft University of Technology\\
Delft, The Netherlands\\
\\
$^3$Philips Research Hamburg\\
Hamburg, Germany\\
\\
\textbf{Running title:} Accelerating Reconstructions Using a Circulant Preconditioner\\


%
%
%
%
\fancypagestyle{title}{%
  \setlength{\headheight}{22pt}%
  \fancyhf{}
  \renewcommand{\headrulewidth}{0pt}
  \renewcommand{\footrulewidth}{0pt}
  \fancyfoot[C]{\thepage}
 \fancyhead[C]{\LARGE \underline{\textbf{Submitted to Magnetic Resonance in Medicine}}}
}%
\thispagestyle{title}
\pagestyle{fancy}
\lhead{}
\rhead{}
\renewcommand{\headrulewidth}{0.0pt}
\renewcommand{\footrulewidth}{0.0pt}

\newpage
\begin{abstract}
\textbf{Purpose}: Design of a preconditioner for fast and efficient parallel imaging and compressed sensing reconstructions.
\newline \textbf{Theory}: Parallel imaging and compressed sensing reconstructions become time consuming when the problem size or the number of coils is large, due to the large linear system of equations that has to be solved in $\ell_1$ and $\ell_2$-norm based reconstruction algorithms. Such linear systems can be solved efficiently using effective preconditioning techniques.
\newline \textbf{Methods}: In this paper we construct such a preconditioner by approximating the system matrix of the linear system, which comprises the data fidelity and includes total variation and wavelet regularization, by a matrix with the assumption that is a block circulant matrix with circulant blocks. Due to its circulant structure, the preconditioner can be constructed quickly and its inverse can be evaluated fast using only two fast Fourier transformations. We test the performance of the preconditioner for the conjugate gradient method as the linear solver, integrated into the Split Bregman algorithm. 
\newline \textbf{Results}: The designed circulant preconditioner reduces the number of iterations required in the conjugate gradient method by almost a factor of~5. The speed up results in a total acceleration factor of approximately~2.5 for the entire reconstruction algorithm when implemented in MATLAB, while the initialization time of the preconditioner is negligible.
\newline \textbf{Conclusion}: The proposed preconditioner reduces the reconstruction time for parallel imaging and compressed sensing in a Split Bregman implementation and can easily handle large systems since it is Fourier-based, allowing for efficient computations.  

\end{abstract}

\newpage

%
%
\section*{Introduction}

The undersampling factor in Parallel Imaging (PI) is in theory limited by the number of coil channels~\cite{Blaimer2004,Deshmane2012,Pruessmann1999,Griswold2002}. 
Higher factors can be achieved by using Compressed Sensing (CS) which estimates missing information by adding a priori information~\cite{Donoho2006,Lustig2007}. The a priori knowledge relies on the sparsity of the image in a certain transform domain. It is possible to combine PI and CS as is done in e.g.~\cite{Otazo2010} and~\cite{Liang2009}, achieving almost an order of magnitude speed-up factors in cardiac perfusion MRI and enabling free-breathing MRI of the liver~\cite{Chandarana2013}. 

\bigskip CS allows for reconstructing an estimate of the true image even in case of considerable undersampling factors, for which the data model generally describes an ill-posed problem without a unique solution. This implies that the true image cannot be found by directly applying Fourier transforms. Instead, regularization is used to solve the ill-posed problem by putting additional constraints on the solution. For CS, such a constraint enforces sparsity of the image in a certain domain, which is promoted by the $\ell_0$-norm~\cite{Boyd2004,Candes2008,Lustig2007}. However, the $\ell_1$-norm is used instead as it is the closest representation that is numerically feasible to implement. The wavelet transform and derivative operators, integrated in total variation regularization, are examples of sparsifying transforms that can be used in the spatial direction \cite{Liang2009,Block1999,Vasanawala2011,Goldstein2009,Murphy2012,Ma2008} and temporal direction \cite{Chandarana2013}, respectively.

\bigskip Although CS has led to a considerable reduction in acquisition times either in parallel imaging applications or in single coil acquisitions, the benefit of the $\ell_1$-norm regularization constraint comes with the additional burden of increased reconstruction times, because $\ell_1$-norm minimization problems are in general difficult to solve. Many methods have been proposed that solve the problem iteratively~\cite{Block1999,Ramani2010,Liu2009,Goldstein2009,Kim2007,Candes2005,Daubechies2004,Donoho2006b}. In this work, we focus on the Split Bregman (SB) approach because of its computational performance~\cite{Bregman1967,Goldstein2009}. SB transforms the initial minimization problem, containing both $\ell_1$ and $\ell_2$-norm terms, into a set of subproblems that either require solving an $\ell_2$-norm minimization problem or an $\ell_1$-norm minimization problem, each of which can be approached using standard methods.

\bigskip The most expensive step in SB, which is also present in many other methods, is to solve an $\ell_2$-norm minimization problem, which can be formulated as a linear least squares problem, e.g.~\cite{Zhou2015}. The system matrix of the least squares problem remains constant throughout the SB iterations and this feature has shown to be convenient for finding an approximation of the inverse system matrix as is done in e.g.~\cite{Cauley2015}. This approach eliminates the need for an iterative scheme to solve the $\ell_2$-norm minimization problem, but for large problem sizes the initial computational costs are high, making it less profitable in practice. 

\bigskip Alternatively, preconditioners can be used to reduce the number of iterations required for solving the least squares problem \cite{Benzi2002}. The incomplete Cholesky factorization and hierarchically structured matrices are examples of preconditioners that reduce the number of iterations drastically in many applications~\cite{Scott2014,Baumann2016}. The drawback of these type of preconditioners is that the full system matrix needs to be built entirely before the reconstruction starts, which for larger problem sizes can only be done on a very powerful computer due to memory limitations. Although in~\cite{Chen2014,Li2015,Xu2015} a penta-diagonal matrix is constructed as preconditioner, solving such a system is still relatively expensive. In addition, before constructing the preconditioner patient-specific coil sensitivity profiles need to be measured, which often leads to large initialization times.  

\bigskip In this work, we design a Fourier transform-based preconditioner for PI-CS reconstructions that takes the coil sensitivities on a patient-specific basis into account, that has negligible initialization time and which is highly scalable to a large number of unknowns, as is often encountered in MRI.

\section*{Theory}
In this section we will first describe the general parallel imaging and compressed sensing problems. Subsequently, the Split Bregman algorithm, which is used to solve this problem, is explained. Hereafter, we introduce the preconditioner that is used to speed up the PI-CS algorithm and elaborate on its implementation and complexity.

\subsubsection*{Parallel Imaging Reconstruction}
\bigskip 
In parallel imaging with full $k$-space sampling the data is described by the model 
\begin{equation} \nonumber
\sF\sS_i \bx = \by_{\text{full},i} \hspace{1cm} \text{for   } i = 1,...,\uNc
\end{equation}
where the $\by_{\text{full},i} \in \mathbb{C}^{N\times 1}$ are the fully sampled k-space data sets for $i\in\{1,..,\uNc\}$ with $\uNc$ the number of coil channels and $\bx\in \mathbb{C}^{N\times 1}$ is the true image~\cite{Pruessmann1999}. Here, $N=m \cdot n$, where $m$ and $n$ define the image matrix size in the $x$ and $y$-directions, respectively, for a 2D sampling case. Furthermore, $\sS_i\in \mathbb{C}^{N\times N}$ are diagonal matrices representing complex coil sensitivity maps for each channel. Finally, $\sF\in \mathbb{C}^{N\times N}$ is the discrete two-dimensional Fourier transform matrix. In case of undersampling, the data is described by the model 
\begin{equation} \label{ill0}
\sR \sF\sS_i \bx = \by_i \hspace{1cm} \text{for   } i = 1,...,\uNc,
\end{equation}
where $\by_{i} \in \mathbb{C}^{N\times 1}$ are the undersampled k-space data sets for $i\in\{1,..,\uNc\}$ with zeros at non-measured $k$-space locations. The undersampling pattern is specified by the binary diagonal sampling matrix $\sR\in\mathbb{R}^{N\times N}$, so that the undersampled Fourier transform is given by $\sR\sF$. Here it is important to note that $\sR$ reduces the rank of $\sR\sF\sS_i$, which means that solving for $\bx$ in Eq.~\eqref{ill0} is in general an ill-posed problem for each coil and a unique solution does not exist. However, if the individual coil data sets are combined and the undersampling factor does not exceed the number of coil channels, the image $\bx$ can in theory be reconstructed by finding the least squares solution, i.e. by minimizing
\begin{equation}\label{ill}
\hat{\bx}= \underset{\bx}{\text{argmin}} \left\{\sum_{i=1}^\uNc \norm{\sR\sF\sS_i \bx -\by_i}_2^2 \right\},
\end{equation}
where $\hat{\bx}\in \mathbb{C}^{N\times 1}$ is an estimate of the true image. 

\subsubsection*{Parallel Imaging Reconstruction with Compressed Sensing}
In case of larger undersampling factors, the problem of solving Eq.~\eqref{ill} becomes ill-posed and additional regularization terms need to be introduced to transform the problem into a well-posed problem. Since MRI images are known to be sparse in some domains, adding $\ell_1$-norm terms is a suitable choice for regularization. The techniques of parallel imaging and compressed sensing are then combined in the following minimization problem
\begin{equation} \label{original}
\hat{\bx} = \underset{\bx}{\text{argmin}} \left\{ \frac{\mu}{2} \sum_{i=1}^{\uNc}\norm{\sR\sF\sS_i \bx - \by_i  }_2^2 + \frac{\lambda}{2} \left( \norm{\sD_x \bx}_1 +\norm{\sD_y \bx}_1 \right) + \frac{\gamma}{2} \norm{ \sW \bx }_1 \right\},
\end{equation}
with $\mu, \lambda$ and $\gamma$ the regularization parameters for the data fidelity, the total variation, and the wavelet, respectively~\cite{Liang2009}. A total variation regularization constraint is introduced by the first-order derivative matrices $\sD_x$, $\sD_y$ $\in \mathbb{R}^{N\times N}$, representing the numerical finite difference scheme
\begin{eqnarray} \nonumber
\left. D_x (x)\right|_{i,j} = x_{i,j} - x_{i-1,j} \hspace{2cm} i=2,..,m,\hspace{2mm} j = 1,..,n \\ \nonumber
\left. D_y (x)\right|_{i,j} = x_{i,j} - x_{i,j-1} \hspace{2cm} i=1,..,m,\hspace{2mm} j = 2,..,n 
\end{eqnarray}
with periodic boundary conditions
\begin{eqnarray}\nonumber
\left. D_x (x)\right|_{1,j} = x_{1,j} - x_{m,j}\hspace{2cm} j = 1,..,n \\ \nonumber
\left. D_y (x)\right|_{i,1} = x_{i,1} - x_{i,n}\hspace{2cm} i=1,..,m
\end{eqnarray}
so that $\sD_x$ and $\sD_y$ are circulant. 
A unitary wavelet transform $\sW\in\mathbb{R}^{N\times N}$ further promotes sparsity of the image in the wavelet domain.

\subsubsection*{Split Bregman Iterations}
Solving Eq.~\eqref{original} is not straightforward as the partial derivatives of the $\ell_1$-norm terms are not well-defined around~0. Instead, the problem is transformed into one that can be solved easily. In this work, we use Split Bregman to convert Eq.~\eqref{original} into multiple minimization problems in which the $\ell_1$-norm terms have been decoupled from the $\ell_2$-norm term, as discussed in detail in~\cite{Bregman1967,Goldstein2009}. For convenience, the Split Bregman method is shown in Algorithm~\ref{euclid}. The Bregman parameters $\bb_x,\bb_y,\bb_w$ are introduced by the Bregman scheme and auxiliary variables $\bd_x,\bd_y,\bd_w$ are introduced by writing the constrained problem as an unconstrained problem. The algorithm consists of two loops: an outer loop and an inner loop. In the inner loop (steps~4-11), we first compute the vector $\bb$ that serves as a right-hand side for step 5, which is solving an $\ell_2$-norm problem. Subsequently, the $\ell_1$-norm subproblems are solved using the shrink function in steps~6-8. Hereafter, the residuals for the regularization terms are computed in steps~9-11 and are subsequently fed back into the system by updating the right hand side vector $\bb$ in step~5. Steps 4-11 can be repeated several times, but one or two inner iterations are normally sufficient for convergence. Similarly, the outer loop feeds the residual encountered in the data fidelity term back into the system, after which the inner loop is executed again. 

\bigskip The system of linear equations, 
\begin{equation}\label{system}
\sA\hat{\bx}=\bb, 
\end{equation}
in line~5 of the algorithm follows from a standard least squares problem, where the system matrix is given by
\begin{equation} \nonumber
\sA= \mu \sum_{i=1}^{\uNc}  \left(\sR\sF \sS_i \right)^H\sR\sF\sS_i  + \lambda  \left( \sD_x^H\sD_x + \sD_y^H\sD_y  \right) + \gamma \sW^H\sW
\end{equation}
with right-hand side 
\begin{equation} \nonumber
\bb= \mu \sum_{i=1}^{\uNc}\left( \sR \sF \sS_i \right)^H\by_i + \lambda \left[\sD_x^H \left( \bd_x^k-\bb_x^k \right) +\sD_y^H \left( \bd_y^k-\bb_y^k \right)  \right] +\gamma \sW^H \left( \bd_w^k - \bb_w^k \right). 
\end{equation} 
In this work we focus on solving Eq.~\eqref{system}, which is computationally the most expensive part of Algorithm~\ref{euclid}. It is important to note that the system matrix $\sA$ remains constant throughout the algorithm and only the right hand side vector $\bb$ changes, which allows us to efficiently solve Eq. \eqref{system}  by using preconditioning techniques. 

\begin{algorithm}[!b]
\caption{Split Bregman Iteration}\label{euclid}
\begin{algorithmic}[1]
\State \textit{Initialize} $\by_i^{[1]}=\by_i \text{ for } i=1,...,\uNc$, $\bx^{[1]}=\text{Sum of Squares}(\by_i, i=1,...,\uNc)$, 
\NoNumber{\textit{Initialize} $\bb_x^{[1]}, \bb_y^{[1]}, \bb_w^{[1]},\bd_x^{[1]},\bd_y^{[1]},\bd_w^{[1]} = \mathbf{0}$}
\State \textbf{for} $j = 1$ to nOuter \textbf{do}
\State \hspace{0.8cm} \textbf{for} $k = 1$ to nInner \textbf{do}
\State \hspace{1.6cm} $\bb = \mu \sum_{i=1}^{\uNc} \sS_i^H \sF^H\sR^H \by_i^{[j]} +  \lambda\left[\sD^H_x(\bd_x^{[k]} - \bb_x^{[k]}) +\sD^H_y(\bd_y^{[k]} - \bb_y^{[k]})  \right] + \gamma\sW^H(\bd_w^{[k]} - \bb_w^{[k]})$
\State \hspace{1.6cm} solve $\sA\bx^{[k+1]}=\bb$ with $\bx^{[k]}$ as initial guess
\State \hspace{1.6cm} $\bd_x^{[k+1]} = \text{shrink}\left(\sD_x\bx^{[k+1]}+\bb_x^{[k]},\frac{1}{\lambda}\right)$
\State \hspace{1.6cm} $\bd_y^{[k+1]} = \text{shrink}\left(\sD_y\bx^{[k+1]}+\bb_y^{[k]},\frac{1}{\lambda}\right)$
\State \hspace{1.6cm} $\bd_w^{[k+1]} = \text{shrink}\left(\sW\bx^{[k+1]}+\bb_w^{[k]},\frac{1}{\gamma}\right)$
\State \hspace{1.6cm} $\bb_x^{[k+1]} = \bb_x^{[k]}+\sD_x\bx^{[k+1]} - \bd_x^{[k+1]} $ 
\State \hspace{1.6cm} $\bb_y^{[k+1]} = \bb_y^{[k]}+\sD_y\bx^{[k+1]} - \bd_y^{[k+1]} $
\State \hspace{1.6cm} $\bb_w^{[k+1]} = \bb_w^{[k]}+\sW\bx^{[k+1]} - \bd_w^{[k+1]}$
\State \hspace{0.8cm} \textbf{end for}
\State \hspace{0.8cm} \textbf{for} $i = 1$ to $\uNc$ \textbf{do}
\State \hspace{0.8cm} \hspace{1cm} $\by_i^{[j+1]} = \by_i^{[j]}+\by_i^{[1]} - \sR\sF\sS_i\bx^{[k+1]}$
\State \hspace{0.8cm} \textbf{end for}
\State \textbf{end for}
\end{algorithmic}
\end{algorithm}

\subsubsection*{Structure of the System Matrix $\sA$}
The orthogonal wavelet transform is unitary, so that $\sW^H\sW=\sI$. Furthermore, the derivative operators were constructed such that the matrices $\sD_x, \sD_y, \sD_x^H$ and $\sD_y^H$ are block circulant with circulant blocks (BCCB). The product and sum of two BCCB matrices is again BCCB, showing that  $\sD_x^H\sD_x + \sD_y^H\sD_y$ is also BCCB. These type of matrices are diagonalized by the two-dimensional Fourier transformation, i.e.
\begin{align}\nonumber
\sD_1  = \sF \sC \sF^H \hspace{5mm}\text{or}\hspace{5mm} \sD_2  = \sF^H \sC \sF
\end{align}
where $\sC$ is a BCCB matrix and $\sD_1$ and $\sD_2$ are diagonal matrices. This motivates us to write the system matrix $\sA$ in Eq.~\eqref{system} in the form 
\begin{align}
\sA &=  \sF^H\sF \sA \sF^H\sF \nonumber \\
&=\sF^H \sK \sF
\end{align}
with $\sK \in \mathbb{C}^{N\times N}$ given by
\begin{eqnarray} \label{K}
\sK = \mu \underbrace{\sum_{i=1}^{\uNc}\sF \sS_i^H\sF^H\sR^H\sR\sF\sS_i \sF^H}_{\sK_c} + \lambda \underbrace{ \vphantom{\sum_{i=1}^{\uNc}}\sF \left( \sD_x^H\sD_x + \sD_y^H\sD_y  \right)\sF^H}_{\sK_d}+ \gamma \underbrace{\vphantom{\sum_{i=1}^{\uNc}}\sI}_{\sK_w}.
\end{eqnarray}

The term $\sD_x^H\sD_x + \sD_y^H\sD_y$ is BCCB, so that $\sK_d$ in $\sK$ becomes diagonal. If there is no sensitivity encoding, that is $\sS_i = \sI$ $\forall i \in \{1,..,\uNc\}$, the entire $\sK$ matrix becomes diagonal in which case the solution $\hat{\bx}$ can be efficiently found by computing 
\begin{equation}
\hat{\bx}=  \sA^{-1}\bb = \sF^H  \sK^{-1} \sF \bb
\end{equation}
for invertible $\sK$. In practice, Fast Fourier Transforms (FFTs) are used for this step. With sensitivity encoding, we have that $\sS_i \ne \sI$ and $\sS_i^H\sF^H\sR^H\sR\sF\sS_i $ is not BCCB for any $i$, hence matrix $\sK$ is not diagonal. In that case we prefer to solve Eq.~\eqref{system} iteratively, since finding $\sK^{-1}$ is now computationally too expensive. It can be observed that the system matrix $\sA$ is Hermitian and positive definite, which motivates the choice for the conjugate gradient (CG) method as an iterative solver.

\subsubsection*{Preconditioning}
A preconditioner $\sM \in \mathbb{C}^{N\times N}$ can be used to reduce the number of iterations required for CG convergence~\cite{Saad2003}. It should satisfy the conditions 
\begin{enumerate}
\item $\sM^{-1}\sA \approx \sI$ to cluster the eigenvalues of the matrix pair around 1, and
\item determination of $\sM^{-1}$ and its evaluation on a vector should be computationally cheap. 
\end{enumerate}
Ideally, we would like to use a diagonal matrix as the preconditioner as this is computationally inexpensive. For this reason, the Jacobi preconditioner is used in many applications with the diagonal elements from matrix~$\sA$ as the input. However, for the current application of PI and CS the Jacobi preconditioner is not efficient since it does not provide an accurate approximate inverse of the system matrix $\sA$. In this work, we use a different approach and approximate the diagonal from $\sK$ in Eq.~\eqref{K} instead. The motivation behind this approach is that the Fourier matrices in matrix~$\sK$ center a large part of the information contained in $ \sS_i^H\sF^H\sR^H\sR\sF\sS_i$ around the main diagonal of~$\sK$, so that neglecting off-diagonal elements of $\sK$ has less effect than neglecting off-diagonal elements of $\sA$. 

\bigskip For the preconditioner used in this work we approximate $\sA^{-1}$ by
\begin{equation}\label{precon} 
\sM^{-1} = \sF^H\text{diag} \{ \bk \}^{-1} \sF,
\end{equation}
where $\text{diag} \{ \}$  places the elements of its argument on the diagonal of a matrix. Furthermore, vector $\bk$ is the diagonal of matrix $\sK$ and can be written as
\begin{eqnarray} \label{Kdiag}
\bk = \mu\bk_c + \lambda \bk_d + \gamma \bk_w,
\end{eqnarray}
where $\bk_c $, $\bk_d$ and $\bk_w$ are the diagonals of $\sK_c$, $\sK_d$ and $\sK_w$, respectively. Note that $\sK_d$ and $\sK_w$ are diagonal matrices already, so that only $\bk_c$ will result in an approximation of the inverse for the final system matrix $\sA$. 

\subsubsection*{Efficient Implementation of the Preconditioner}
The diagonal elements $\bk_{c;i}$ of $\sK_{c;i}=\underbrace{\sF\sS_i^H\sF^H}_{\sC_i^H}\underbrace{\vphantom{\sF\sS_i^H\sF^H}\sR^H\sR}_{\sR}\underbrace{\sF\sS_i\sF^H \vphantom{\sF\sS_i^H\sF^H}}_{\sC_i}$ for a certain $i$ are found by noting that $\sC_i=\sF\sS_i\sF^H$ is in fact a BCCB matrix. The diagonal elements $\bk_{c;i}$ of $\sK_{c;i}$ can now be found on the diagonal of  $\sC_i^H  \sR \sC_i$, so that 
\begin{align}
\bk_{c;i} = \sum_{j=1}^N \be_j\left( \bc_{j;i}^H \sR \bc_{j;i} \right)\nonumber,
\end{align} 
with $\bc_{j;i}^H$ being the $j^\text{th}$ row of matrix $\sC_i^H$ and $\be_j$ the $j^\text{th}$ standard basis vector. Note that the scalar $\left( \bc_{j;i}^H \sR \bc_{j;i} \right)$ is the $j^\text{th}$ entry of vector $\bk_{c;i}$. Since $\sR$ is a diagonal matrix which can be written as $\sR=\text{diag}\{\br\}$, we can also write
\begin{align} 
\bk_{c;i} &= \sum_{j=1}^N \be_j\left( \bc_{j;i}^H \circ \bc_{j;i}^T \right) \br \nonumber \\
&= 
\begin{bmatrix} 
\bc_{1;i}^H \circ \bc_{1;i}^T \\
\bc_{2;i}^H \circ \bc_{2;i}^T \\
\vdots\\
\bc_{N;i}^H \circ \bc_{N;i}^T
\end{bmatrix}
\br \nonumber\\\label{d}
&=\left( \sC_i^H \circ \sC_i^T \right)\br,
\end{align} 
where $\circ$ denotes the element-wise (Hadamard) product. Since the element-wise product of two BCCB matrices is again a BCCB matrix, the circular convolution theorem tells us~\cite{Bracewell1986,Gray2006} that
\begin{equation} \nonumber
\mathcal{F}\left\{ \bk_{c;i} \right\} = \mathcal{F}\left\{ \left( \bc_{1;i}^H \circ \bc_{1;i}^T \right)^T* \br \right\} =\mathcal{F}\left\{ \left(\bc_{1;i}^H \circ \bc_{1;i}^T\right)^T \right\}\circ \mathcal{F}\left\{ \br \right\}.
\end{equation}
Here, $\mathcal{F}$ denotes the two-dimensional Fourier transform of a vector that is reshaped in matrix form with dimensions $m\times n$. The resulting matrix vector product in Eq.~\eqref{d} can now be efficiently computed as 
\begin{align}
\bk_{c;i} = \mathcal{F}^{-1}\left\{\mathcal{F}\left\{ \left(\bc_{1;i}^H \circ \bc_{1;i}^T \right)^T \right\}\circ\mathcal{F}\left\{\br\right\}\right\}.
\label{eq:diagonalelements}
\end{align}
Finally, the diagonal elements $\bd$ of the diagonal matrix $\sD$ with structure $\sD = \sF \sC \sF^H$ can be computed efficiently by using $\bd=\mathcal{F}\left\{ \bc_1 \right\}$, where $\bc_1$ is the first row of $\sC$. Therefore, the first row $\bc_{1;i}^H$ of matrix $\sC_i^H$ is can be found by $\bc_{1;i}^H = \mathcal{F}^{-1}\left\{ \bs_i^H \right\}$, with $\bs_i^H$ being the diagonal elements of matrix $\sS_i$. For multiple coils Eq.~\eqref{eq:diagonalelements} becomes 
\begin{align} 
\bk_c = \mathcal{F}^{-1}\left\{\mathcal{F}\left\{ \sum_{i=1}^\uNc \left( \bc_{1;i}^H \circ \bc_{1;i}^T\right)^T \right\}\circ\mathcal{F}\left\{\br\right\}\right\}.
\end{align}

\bigskip Since $\sD_x^H\sD_x + \sD_y^H\sD_y$ is BCCB, the elements of $\bk_d$ can be quickly found by evaluating $\bk_d=\mathcal{F}\left\{ \bt_1 \right\}$, where $\bt_1$ is the first row of $\sD_x^H\sD_x + \sD_y^H\sD_y$. Finally, the elements of $\bk_w$ are all equal to one, since $\sK_\omega$ is the identity matrix.

\subsubsection*{Complexity}
For every inner-iteration of the Split Bregman algorithm we need to solve the linear system given in Eq.~\eqref{system}, which is done iteratively using a Preconditioned Conjugate Gradient method (PCG). In this method, the preconditioner constructed above is used as a left preconditioner by solving the following system of equations:
\begin{align}
\sM^{-1}\sA\hat{\bx}=\sM^{-1}\bb,
\end{align}
where $\hat{\bx}$ is the approximate solution constructed by the PCG algorithm. In PCG this implies that for every iteration the preconditioner should be applied once on the residual vector $\br = \sA\hat{\bx}-\bb$. The preconditioner $\sM$ can be constructed beforehand since it remains fixed for the entire Split Bregman algorithm as the parameters $\mu$, $\lambda$, and $\gamma$ are constant. As can be seen in Table~\ref{table_constmevalm}, $\sM^{-1}$ is constructed in $(3+2\uNc)N + (4+\uNc)N\log N$ FLOPS only. Evaluation of the diagonal preconditioner $\sM^{-1}$ from Eq.~\eqref{precon} on a vector amounts to two Fourier transforms and a single multiplication, and therefore requires $N+2N\log N$ FLOPS.  

\begin{figure}[h]
\centering
\includegraphics[scale=0.65]{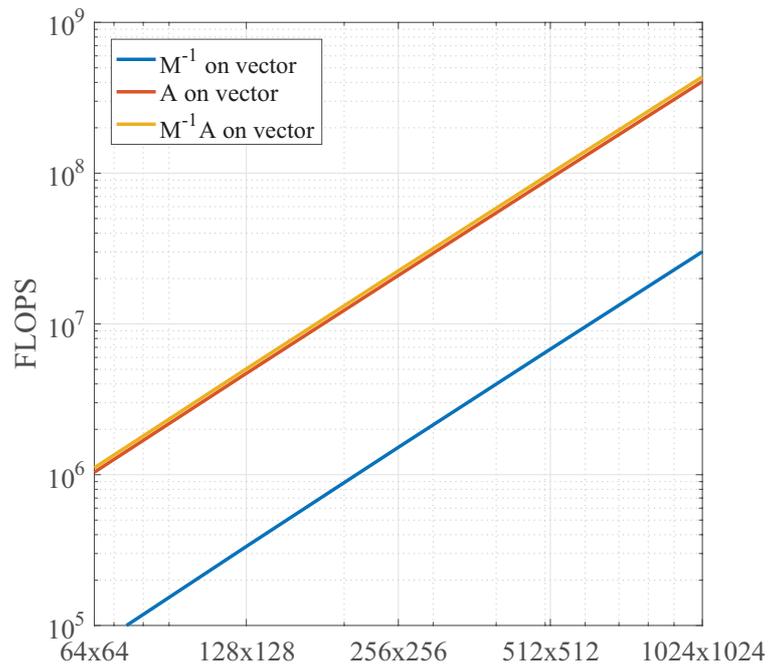}
\caption{The complexity for different problem sizes. The number of flops for the action of the preconditioner $\sM$ on a vector (blue), $\sA$ on a vector (red), and the combination of the two (yellow) are depicted for $\sN_\text{c} = 12$.}
\label{fig:complexity}
\end{figure}

To put this into perspective, evaluation of matrix $\sA$ on a vector requires $(6+4\uNc)N + 2\uNc N\log N$ FLOPS, as shown in Table~\ref{table_constmevalm}. The upper bound on the additional costs per iteration relative to the costs for evaluating $\sA$ on a vector is therefore 
\begin{align}
\lim_{N\to\infty}\frac{N+2N\log N}{(6+4\uNc)N + 2\uNc N\log N} = \frac{1}{\uNc},\nonumber
\end{align}
showing that the preconditioner evaluation step becomes relatively cheaper for an increasing number of coil elements. The scaling of the complexity with respect to the problem size is depicted in Fig.~\ref{fig:complexity} for a fixed number of coils $\text{N}_\text{c}=12$. 

\section*{Methods}

\subsubsection*{MR Data Acquisition}
Two fully sampled data sets were acquired on a healthy volunteer after giving informed consent. The Leiden University Medcal Center Committee for Medical Ethics approved the experiment. An Ingenia 3T dual transmit MR system (Philips Healthcare, Best, The Netherlands) was used to acquire the \textit{in vivo} data. A 12-element posterior receiver array and 15-channel head coil were used for reception in the spine and the brain, respectively, and the body coil was used for RF transmission.

\bigskip For the spine data set, T$_1$-weighted images were acquired using a turbo spin-echo (TSE) sequence with the following parameters: field of view (FOV) =  340$\times$340 mm$^2$; in-plane resolution 0.66$\times$0.66 mm$^2$; slice thickness = 4 mm; slices = 15; echo time (TE)/ repetition time (TR) / TSE factor = 8 ms/ 648 ms/ 8; flip angle (FA) = 90$\degree$; refocusing angle = 120$\degree$; water-fat shift = 1.5 pixels; and scan time = 2:12 min. T$_2$-weighted TSE scans had the following parameters: FOV =  340$\times$340 mm$^2$; in-plane resolution 0.66$\times$0.66 mm$^2$; slice thickness = 4 mm; slices = 15; TE/TR/TSE factor = 113 ms/ 4008 ms/32; FA = 90$\degree$; water-fat shift = 1.1 pixels; and scan time = 3:36 min. 

\bigskip For the brain data set, T$_1$-weighted images were acquired using an inversion recovery turbo spin-echo (IR TSE) sequence with the following parameters: field of view (FOV) =  230$\times$230 mm$^2$; in-plane resolution 0.90$\times$0.90 mm$^2$; slice thickness = 4 mm; slices = 24; echo time (TE)/ repetition time (TR) / TSE factor = 20 ms/ 2000 ms/ 8; refocusing angle = 120$\degree$; IR delay: 800 ms, water-fat shift = 2.6 pixels; and scan time =  05:50 min. 

\subsubsection*{Coil sensitivity maps}
Unprocessed k-space data was stored per channel and used to construct complex coil sensitivity maps for each channel \cite{Uecker2014}. Note that the coil sensitvity maps are normalized such that 
$$\hat{S_i}=\left[\sum_{j=1}^{\uNc}{S_j^H S_j} \right]^{-\frac{1}{2}} S_i \hspace{1cm} \text{for } i=1,...,\uNc.$$ 
The normalized coil sensitivity maps were given zero intensity outside the subject, resulting in an improved SNR of the final reconstructed image. For the data model to be consistent, also the individual coil images were normalized according to 
$$m_i = \hat{S}_i \sum_{j=1}^{\uNc} \hat{S}_j^H m_j \hspace{1cm} \text{for } i=1,...,\uNc.$$

\subsubsection*{Coil Compression}
To study the effect of coil compression on the performance of the constructed preconditioner, reconstruction for the spine data set was performed with and without coil compression. A compression matrix was constructed as done in \cite{Huang2008}, and multiplied by the normalized individual coil images and the coil sensitivity maps, to obtain virtual coil images and its corresponding virtual coil sensitivity maps. The six least dominant virtual coils were ignored to speed up the reconstruction, while satisfying SNR was obtained.

\subsubsection*{Undersampling}
\begin{figure*}[h]
\centering
\includegraphics[width=0.8\linewidth]{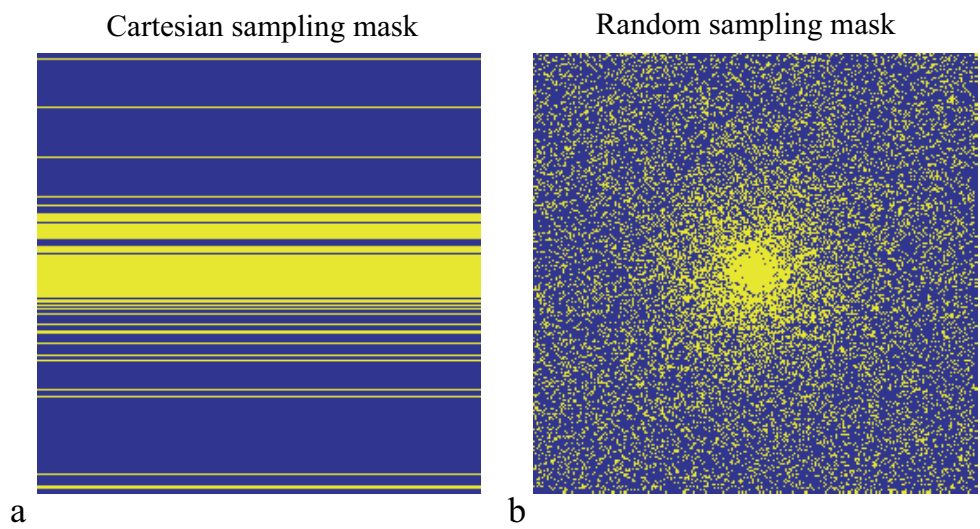}
\caption{k-Space subsampling patterns used. The Cartesian sampling mask (a) has a variable density pattern with undersampling in the foot-head direction. The random sampling mask with variable density (b) is used to show the generalizability of the approach to other sampling schemes. For both masks an undersampling factor of 4 is used.}
\label{fig:Masks}
\end{figure*}

Two undersampling schemes are studied: a random line pattern with variable density in the foot-head direction and a fully random pattern with variable density, as shown in Fig.~\ref{fig:Masks}. Undersampling factors of four (R=4) and eight (R=8) were studied.  

\subsubsection*{Reconstruction}
The Split Bregman algorithm was implemented in MATLAB (The MathWorks, Inc., Natick, MA, USA). All image reconstructions were performed on a Windows 64-bit machine with an Intel i3-4160 CPU @ 3.6 GHz and 8 GB internal memory.

\bigskip Reconstructions were performed for reconstruction matrix sizes of $128\times 128$, $256\times 256$, and $512\times 512$, and the largest reconstruction matrix was interpolated to obtain a simulated data set of size $1024 \times 1024$ for theoretical comparison. To investigate the effect of the regularization parameters on the performance of the preconditioner, three different regularization parameter sets were chosen as:
\begin{enumerate}
\item set 1  $\mu=10^{-3}$,  $\lambda=4 \cdot 10^{-3}$, and $\gamma=10^{-3}$
\item set 2  $\mu=10^{-2}$,  $\lambda=4 \cdot 10^{-3}$, and $\gamma=10^{-3}$
\item set 3  $\mu=10^{-3}$,  $\lambda=4 \cdot 10^{-3}$, and $\gamma=4\cdot10^{-3}.$
\end{enumerate}
The Daubechies~4 wavelet was used. Furthermore, the Split Bregman algorithm was performed with an inner loop of one iteration and an outer loop of 20 iterations. The tolerance (relative residual norm) in the PCG algorithm was set to $\varepsilon = 10^{-3}$. 

\section*{Results}

\begin{figure}[h]
\centering
\includegraphics[width=1\linewidth]{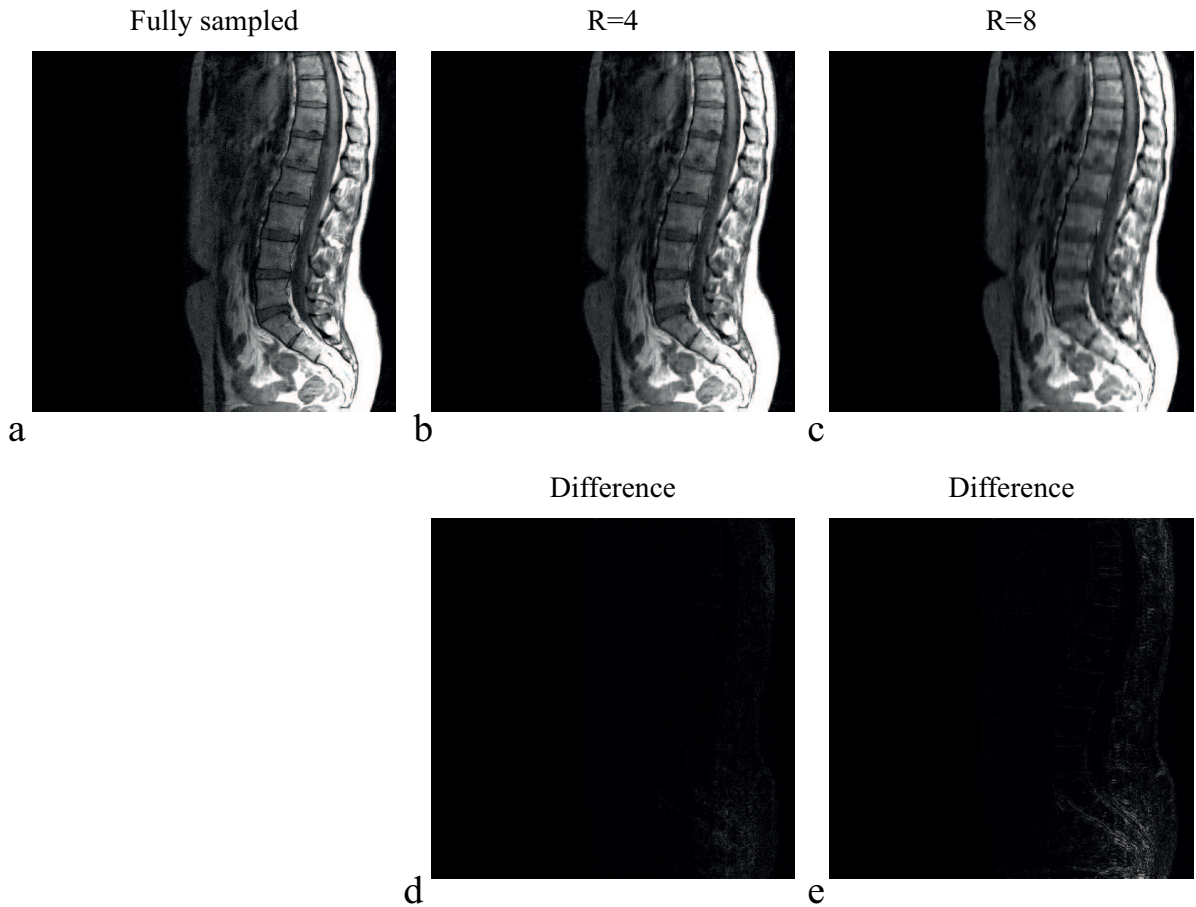}
\caption{Reconstruction results for different Cartesian undersampling factors. (a) shows the fully sampled scan as a reference, whereas (b) and (c) depict the reconstruction results for undersampling factors four (R=4) and eight (R=8), respectively. The absolute difference is shown in (d) and (e) for R=4 and R=8, respectively. The reconstruction matrix has dimensions $512\times512$. Regularization parameters where set to $\mu = 1\cdot 10^{-3}, \lambda =4\cdot 10^{-3}$, and $\gamma = 1\cdot 10^{-3}$.}
\label{fig:recons}
\end{figure}

Figure~\ref{fig:recons} shows the T$_1$-weighted TSE spine images for a reconstruction matrix size of $512\times 512$, reconstructed with the SB implementation for a fully sampled data set and for undersampling factors of four (R=4) and eight (R=8), where Cartesian sampling masks were used. The quality of the reconstructed images for R=4 and R=8 demonstrate the performance of the compressed sensing algorithm. The difference between the fully sampled and undersampled reconstructed images are shown in Fig.~\ref{fig:recons}d and Fig.~\ref{fig:recons}e for R=4 and R=8, respectively.

\begin{figure*}[h]
\centering
\includegraphics[width=\linewidth]{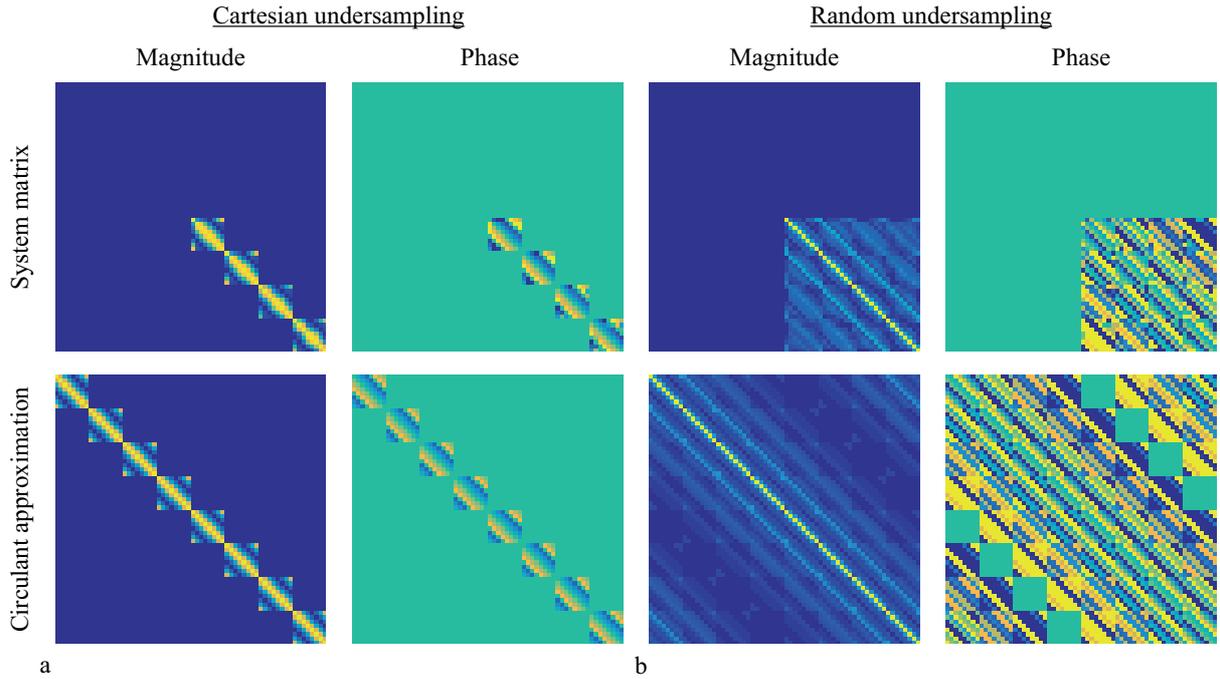}
\caption{System matrix and circulant approximated system matrix. The first two columns and the last two columns show the system matrix elements for Cartesian (a) and random (b) undersampling, respectively. The top row depicts the elementwise magnitude and phase for the true system matrix $\sA$, whereas the bottom row depicts the elementwise circulant approximated system matrix.}
\label{fig:MatrixAandK}
\end{figure*}

\bigskip The fully built system matrix $\sA = \sF^H  \sK \sF$ is compared with its circulant approximation $\sF^H \text{diag}\{\bk \} \sF$ in Fig.~\ref{fig:MatrixAandK}a for Cartesian undersampling and a system matrix size of $64 \times 64$. Both the magnitude and phase parts of $\sA$ contain many zeros due to the lack of coil sensitivity in a large part of the image domain when using the posterior coil. These zeros are not present in the circulant approximation, since the circulant property is enforced by neglecting all off-diagonal elements in $\sK$. The introduced entries in the circulant approximation do not add relevant information to the system, because the image vector on which the system matrix acts contains zero signal in the region corresponding with the newly introduced entries. Therefore, the magnitude and phase are approximated well by assuming the circulant property. Fig.~\ref{fig:MatrixAandK}b shows the same results for random undersampling, demonstrating the generalisability of this approach to different sampling schemes.    
 
\bigskip Table \ref{inittime} reports the number of seconds needed to build the circulant preconditioner in MATLAB before the reconstruction starts, for different orders of the reconstruction matrix. Note that the actual number of unknowns in the corresponding systems is equal to the number of elements in the reconstruction matrix size, which leads to more than 1 million unknowns for the $1024\times1024$ case. For all matrix sizes the initialization time is negligible compared with the image reconstruction time. 

\begin{figure}[h]
\centering
\includegraphics[width=1\linewidth]{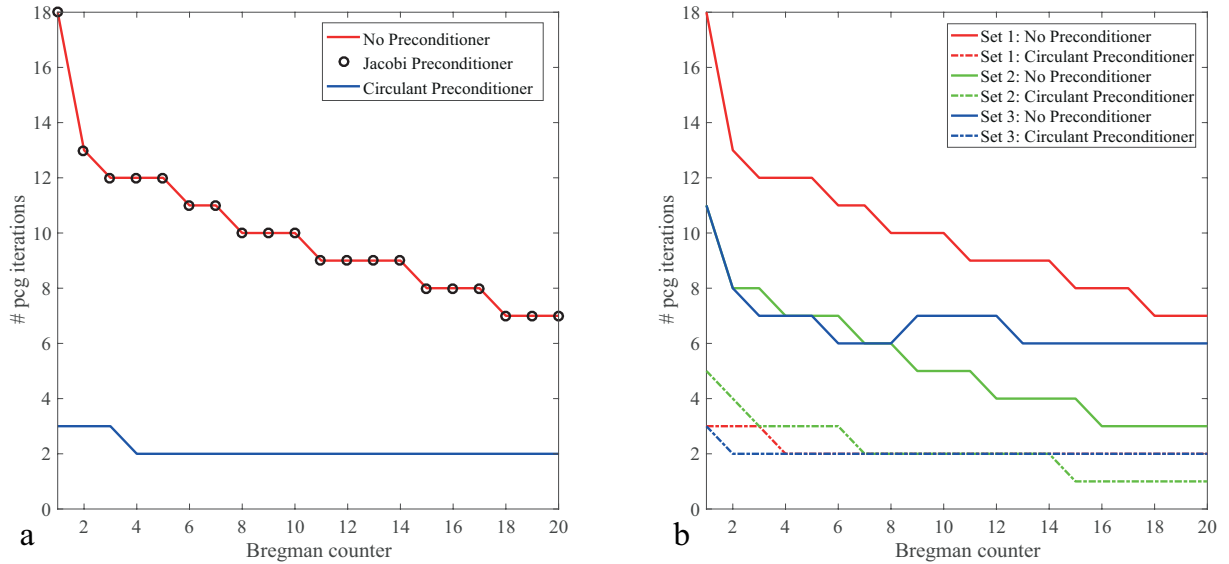}
\caption{Number of iterations needed per Bregman iteration. The circulant preconditioner reduces the number of iterations considerably compared with the non-preconditioned case. The Jacobi preconditioner does not reduce the number of iterations due to the poor approximation of the system matrix' inverse. (a) depicts the iterations for Set 1: $\left( \mu = 1\cdot 10^{-3}, \lambda =4\cdot 10^{-3}, \gamma = 1\cdot 10^{-3} \right)$, whereas (b) depicts the iterations for Set 1, Set 2: $\left(\mu =1\cdot 10^{-2}, \lambda =4\cdot 10^{-3}, \gamma = 1\cdot 10^{-3}\right)$, and Set 3: $\left(\mu = 1\cdot 10^{-3}, \lambda =4\cdot 10^{-3},  \gamma = 4\cdot 10^{-3}\right)$ The preconditioner shows the largest speed up factor when the regularization parameters are well-balanced.}
\label{fig:iterations}
\end{figure}

\bigskip Figure \ref{fig:iterations}a shows the number of iterations required for PCG to converge in each Bregman iteration without preconditioner (red line), with the Jacobi preconditioner (black circles) and with the circulant preconditioner (blue line) for regularization parameters  $\mu=10^{-3}, \lambda = 4 \cdot 10^{-3}$ and $\gamma = 10^{-3}$ and a reconstruction matrix size of $256 \times 256$. The Jacobi preconditioner does not reduce the number of iterations, which shows that the diagonal of the system matrix $\sA$ does not contain enough information to result in a good approximation of $\sA^{-1}$. Moreover, it shows that the linear system is invariant under scaling. The circulant preconditioner, however, reduces the number of iterations considerably, with a reduction factor of 6 for the first Bregman iteration and a reduction factor of 3.5 for the last Bregman iteration, leading to a total speed-up factor of 4.65 in the PCG part.         

\begin{figure}[h]
\centering
\includegraphics[width=1\linewidth]{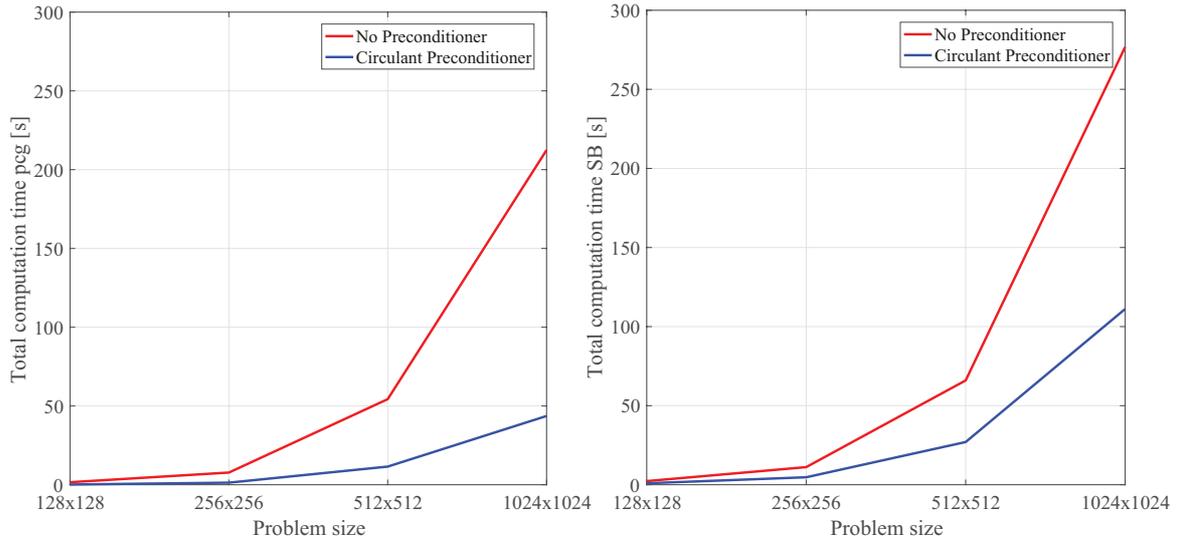}
\caption{Computation time for 20 Bregman iterations and different problem sizes. (a) Using the preconditioner, the total computation time for the PCG part in 20 Bregman iterations is reduced by more than a factor of 4.5 for all studied problem sizes. (b) The computation time for 20 Bregman iterations of the entire algorithm also includes the Bregman update steps, so that the total speedup factor is approximately 2.5 for the considered problem sizes.}
\label{fig:computationtime}
\end{figure}

\bigskip The effect of the reduced number of PCG iterations can directly be seen in the computation time for the reconstruction algorithm, plotted in Fig.~\ref{fig:computationtime} for different problem sizes. Figure \ref{fig:computationtime}a shows the total PCG computation time when completing the total Split Bregman method, whereas Fig. \ref{fig:computationtime}b shows the total computation time required to complete the entire reconstruction algorithm. A fivefold gain is achieved in the PCG part by reducing the number of PCG iterations, which directly relates to the results shown in Fig.~\ref{fig:iterations}a. The overall gain of the complete algorithm, however, is a factor 2.5 instead of 5, which can be explained by the computational costs of the update steps outside the PCG iteration loop as described in Algorithm~\ref{euclid}. 

\bigskip The performance of the constructed preconditioner is compared for three different regularization parameters, as listed in the Method section. The number of iterations required by PCG for each Bregman iteration is shown in Fig.~\ref{fig:iterations}b for the three studied parameter sets. The preconditioned case (dotted line) always outperforms the non-preconditioned case (solid line), but the speed up factor depends on the regularization parameters. Parameter set 1 depicts the same result as shown in Fig.~\ref{fig:iterations}a and results in the best reconstruction of the fully sampled reference image. In parameter set 2 more weight is given to the data fidelity term by increasing the parameter $\mu$. Since the preconditioner relies on an approximation of the data fidelity term, it performs less optimally than for relatively smaller $\mu$ (such as in set 1) for the first few Bregman iterations, but there is still a threefold gain in performance. Finally, there is hardly any change for parameter set 3 compared with parameter set 1, because the larger wavelet regularization parameter $\gamma$ gives more weight to a term that was integrated in the preconditioner in an exact way, as for the total variation term, without any approximations.

\begin{figure}[h]
\centering
\includegraphics[width=0.7\linewidth]{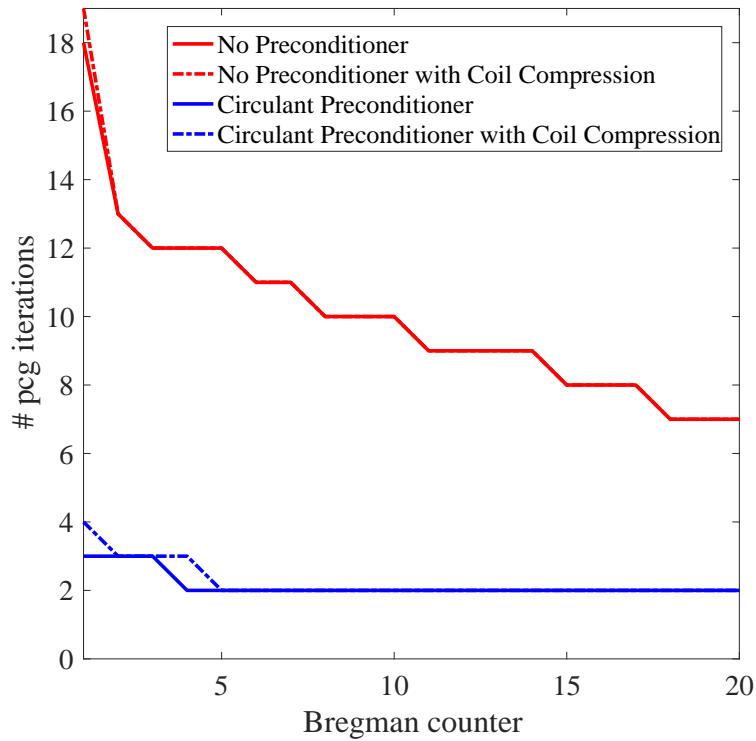}
\caption{The effect of coil compression. Shown are the number of iterations needed per Bregman iteration with and without coil compression applied. The solid lines and the dashed lines depict the results with and without coil compression, respectively.}
\label{fig:coilcompression}
\end{figure}

\bigskip Coil compression can be applied to reduce the reconstruction time. The effect on the number of iterations required for PCG when applying coil compression on the measurement data is expected to be the same, as it only slightly affects the structure or content of the system matrix $\sA$. Figure~\ref{fig:coilcompression} illustrates the result on the required iterations when half of the coils are taken into account. Only a small discrepancy is encountered for the first few iterations, which demonstrates that coil compression and preconditioning can be combined to optimally reduce the reconstruction time. 

\begin{figure}[!t]
\centering
\includegraphics[width=0.7\linewidth]{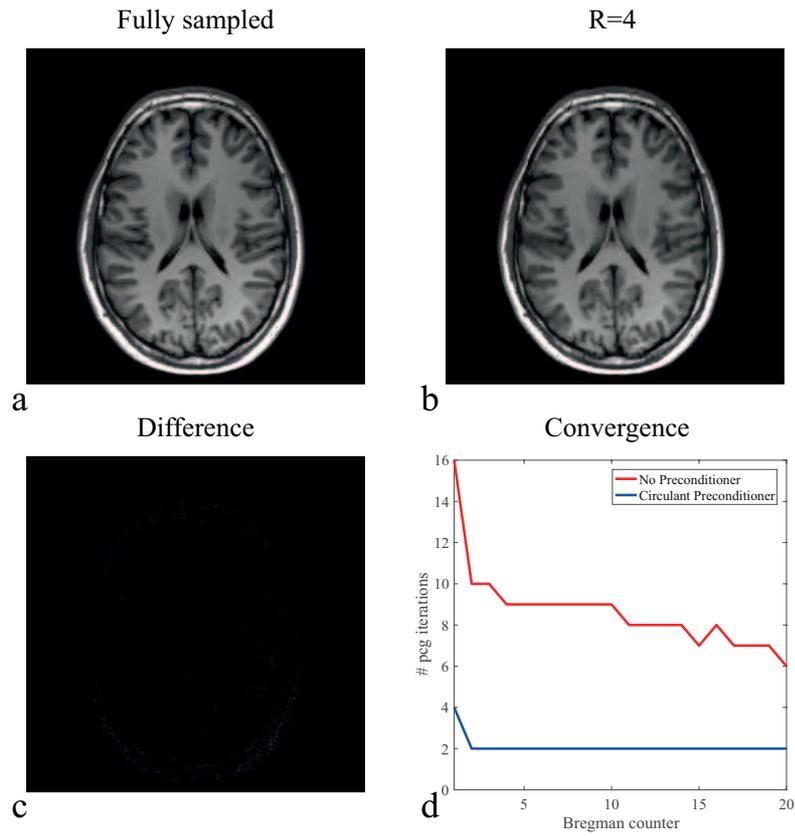}
\caption{Reconstruction results for a brain scan. (a) shows the fully sampled scan as a reference, whereas (b) depicts the reconstruction results for an undersampling factor of four (R=4). The absolute difference is shown in (c). The reconstruction matrix has dimensions $256\times256$ and regularization parameters where chosen as $\mu = 1\cdot 10^{-3}, \lambda =4\cdot 10^{-3}$, and $\gamma = 2\cdot 10^{-3}$. The convergence results for the PCG part with and without preconditioner are plotted in (d), showing similar reduction factors as with the posterior coil.}
\label{fig:reconsbrain}
\end{figure}

\bigskip The method also works for different coil configurations. In Fig.~\ref{fig:reconsbrain} the result is shown when using the 15-channel head coil for a brain scan. The circulant preconditioner clearly reduces the number of iterations, with an overall speed-up factor of~4.1 in the PCG part.

\section*{Discussion}
In this work we have introduced a preconditioner that reduces the reconstruction times for CS and PI problems. The Split Bregman algorithm has been used to solve the corresponding minimization problem, in which the most time-consuming step is solving an $\ell_2$-norm minimization problem. This $\ell_2$-norm minimization problem is written as a linear system of equations characterized by the system matrix~$\sA$ in Eq.~\eqref{system}. The effectiveness of the introduced preconditioner comes from the fact that the system matrix is approximated as a BCCB matrix. Both the total variation and the wavelet regularization terms are BCCB, which means that only the data fidelity term, which is not BCCB due to the sensitivity profiles of the receive coils and the undersampling of k-space, is approximated by assuming a BCCB structure in the construction of the preconditioner. This approximation was shown to be accurate for CS-PI problem formulations. The efficiency of this approach comes from the fact that BCCB matrices are diagonalized by Fourier transformations, which means that the inverse of the preconditioner can simply be found by inverting a diagonal matrix and applying two additional FFTs. 

\bigskip The designed preconditioner allows for solving the most expensive $\ell_2$-norm problem almost 5 times faster than without preconditioning, resulting in an overall speed up factor of about~2.5. The discrepancy between the two speed up factors can be explained by the fact that apart from solving the linear problem, also remaining update steps need to be performed. Especially step~4 and step~13-15 of Algorithm 1 are time consuming as for each coil a 2D Fourier transform needs to be performed. Furthermore, the wavelet computation in step~4, 8, and 11 are time consuming factors as well. Therefore, speed up factors higher than 2.5 are expected for an optimized Bregman algorithm. Further acceleration can be obtained through coil compression~\cite{Huang2008,Zhang2013}, as the results in this study showed that it has negligible effect on the performance of the preconditioner.

\bigskip The time required to construct the preconditioner is negligible compared with the reconstruction times as it involves only a few FFTs. The additional costs of applying the preconditioner on a vector is negligible as well, because it involves only two Fourier transformations and an inexpensive multiplication with a diagonal matrix. Therefore, the method is highly scalable and can handle large problem sizes.

\bigskip The preconditioner works optimally when the regularization terms in the minimization problem are BCCB matrices in the final system matrix. This implies that the total variation operators should be chosen such that the final total variation matrix is BCCB, and that the wavelet transform should be unitary. However, both the system matrix and the preconditioner can be easily adjusted to support single regularization instead of the combination of two regularization approaches, as was implemented in this study. 

\bigskip The BCCB approximation for the data fidelity term supports both Cartesian and random undersampling patterns and works well for different undersampling factors. Furthermore, it takes into account each receiver coil, as was shown to perform well both for the 12-channel posterior coil and the 15-channel head coil. Hence, the method is flexible to cope with a variety of MR configurations. 

\bigskip This work focussed on the linear part of the Split Bregman method, in which only the right-hand side vector changes in each iteration and not the system matrix. Other $\ell_1$-norm minimization algorithms exist that require a linear solver~\cite{Rodriguez2009}, such as IRLS or Second-Order Cone Programming. For those type of algorithms linear preconditioning techniques can be applied as well. Although the actual choice for the preconditioner depends on the system matrix of the linear problem, which is in general different for different minimization algorithms, similar techniques as used in the current work can be exploited to construct a preconditioner for other minimization algorithms that involve linear solvers. 

\bigskip The regularization parameters were shown to influence the performance of the preconditioner. Since the only approximation in the preconditioner comes from the approximation of the data fidelity term, the preconditioner results in poorer performance if the data fidelity term is very large compared with the regularization terms. In practice, such a situation is not likely to occur if the regularization parameters are chosen such that an optimal image quality is obtained in the reconstructed image.  

\bigskip Future work will focuss on the implementation of the preconditioned Split Bregman algorithm in 3D imaging applications. 
%
\section*{Acknowledgment}
We would like to thank Ad Moerland from Philips Healthcare Best (The Netherlands) and Mariya Doneva from Philips Research Hamburg (Germany) for helpful discussions on reconstruction.

%
%

\newpage
\section*{Tables}

\begin{table}[h]
\renewcommand{\arraystretch}{1.3}
\caption{FLOPS required for construction of $\sM^{-1}$ and for evaluation of $\sA$ on a vector}
\label{table_constmevalm}
\centering
\begin{tabular}{c|l|c|}
\cline{2-3}
& \bfseries Operation &\bfseries FLOPS \\ \cline{1-3}
\multicolumn{1}{| c | }{\multirow{7}{*}{Construction of $\sM^{-1}$} } & $\bc_i^H = \mathcal{F}^{-1}\left\{ \bs_i^H \right\}$ $\forall i \in \{1,..,\uNc\}$, & $\uNc N\log N$\\  
\multicolumn{1}{| c | }  {} &$\sum_i^\uNc \left( \bc_{1;i}^H \circ \bc_{1;i}^T\right)^T $ & $2 \uNc N - N$ \\  
\multicolumn{1}{| c | }  {} &$\mathcal{F}^{-1}\left\{\mathcal{F}\left\{ \hdots \right\}\circ\mathcal{F}\left\{\hdots\right\}\right\} $ & $N + 3N\log N$\\  
\multicolumn{1}{| c | }  {} & $\bk_{d} = \mathcal{F}^{-1}\left\{ \bt_1 \right\}$  & $N\log N$ \\  
\multicolumn{1}{| c | }  {} & $\bk = \bk_c + \bk_{d} + \bk_w$ & $2N$ \\
\multicolumn{1}{| c | }  {} & $\bk^{-1}$ & $N$ \\  \cline{2-3}
\multicolumn{1}{| c | }  {} & $\textbf{Total}$ & $(3+2\uNc)N + (4+\uNc)N\log N$\\ \cline{1-3}
\multicolumn{1}{| c | }{\multirow{5}{*}{Evaluation $\sA$ on vector}  } & $\sum_{i=1}^{\uNc}  \left(\sR\sF \sS_i \right)^H\sR\sF\sS_i $  & $\uNc (3N + 2N\log N) +\uNc N-N$ \\  
\multicolumn{1}{| c | }  {} & $\sD_x^H\sD_x + \sD_y^H\sD_y $ & $5N$ \\  
\multicolumn{1}{| c | }  {} & $\sW^H\sW$ & $0$ \\
\multicolumn{1}{| c | }  {} & Summation of the three terms above & $2N$\\   \cline{2-3}
\multicolumn{1}{| c | }  {} & $\textbf{Total}$ & $(6+4\uNc)N + 2\uNc N\log N$ \\ \cline{1-3}
\end{tabular}
\end{table}

\begin{table}[h]
\caption{Initialization times for constructing the preconditioner for different problem sizes. Even for very large problem sizes the initialization time does not exceed two seconds. Additional costs are given as percentage of the total reconstruction time without preconditioning.}
\begin{center}
\begin{tabular}{| l || c  | c | c | r |}  \hline
Problem size        & $128\times128$ & $256\times256$  & $512\times512$   & $1024\times1024$ \\ \hline
Initialization time (s) & 0.0395 & 0.0951  & 0.3460 & 1.3371 \\ \hline
Additional costs (\%) & 1.7 & 0.85 & 0.52 & 0.48 \\ \hline
\end{tabular}
\end{center}\label{inittime}
\end{table}

\newpage
\section*{Figures}

\end{document}